\documentclass[letterpaper]{article} 
\usepackage{aaai24}  
\usepackage{times}  
\usepackage{helvet}  
\usepackage{courier}  
\usepackage[hyphens]{url}  
\usepackage{graphicx} 
\usepackage{natbib}  
\usepackage{caption} 
\usepackage{algorithm}
\usepackage{algorithmic}

\usepackage{newfloat}
\usepackage{listings}
\DeclareCaptionStyle{ruled}{labelfont=normalfont,labelsep=colon,strut=off} 
\floatstyle{ruled}
\newfloat{listing}{tb}{lst}{}
\floatname{listing}{Listing}
\usepackage{amsfonts}
\usepackage{amsmath}
\usepackage{amssymb}
\usepackage{multirow}
\usepackage{tabularx}
\usepackage{booktabs}
\usepackage{adjustbox}
\newcommand*\rot{\rotatebox{90}}
\usepackage[accsupp]{axessibility}
\usepackage{enumitem}

\usepackage{xcolor,colortbl}
\definecolor{cadmiumred}{rgb}{0.89, 0.0, 0.13}

\title{Expediting Contrastive Language-Image Pretraining\\via Self-distilled Encoders}
\author{
    Bumsoo Kim\thanks{correspondence to: bumsoo.kim@lgresearch.ai},
    Jinhyung Kim,
    Yeonsik Jo,
    Seung Hwan Kim
}
\affiliations{
    LG AI Research\\
}

\begin{document}

\maketitle

\begin{abstract}
   Recent advances in vision language pretraining (VLP) have been largely attributed to the large-scale data collected from the web.
   However, uncurated dataset contains weakly correlated image-text pairs, causing data inefficiency.
   To address the issue, knowledge distillation have been explored at the expense of extra image and text momentum encoders to generate teaching signals for misaligned image-text pairs.
   In this paper, our goal is to resolve the misalignment problem with an efficient distillation framework.
   To this end, we propose ECLIPSE: Expediting Contrastive Language-Image Pretraining with Self-distilled Encoders.
   ECLIPSE features a distinctive distillation architecture wherein a shared text encoder is utilized between an online image encoder and a momentum image encoder.
   This strategic design choice enables the distillation to operate within a unified projected space of text embedding, resulting in better performance.
   Based on the unified text embedding space, ECLIPSE compensates for the additional computational cost of the momentum image encoder by expediting the online image encoder.
   Through our extensive experiments, we validate that there is a sweet spot between expedition and distillation where the partial view from the expedited online image encoder interacts complementarily with the momentum teacher.
   As a result, ECLIPSE outperforms its counterparts while achieving substantial acceleration in inference speed.
\end{abstract}
\section{Introduction}
\label{sec:introduction}
\begin{figure}[t]
    \centering
    \includegraphics[width=0.9\columnwidth]{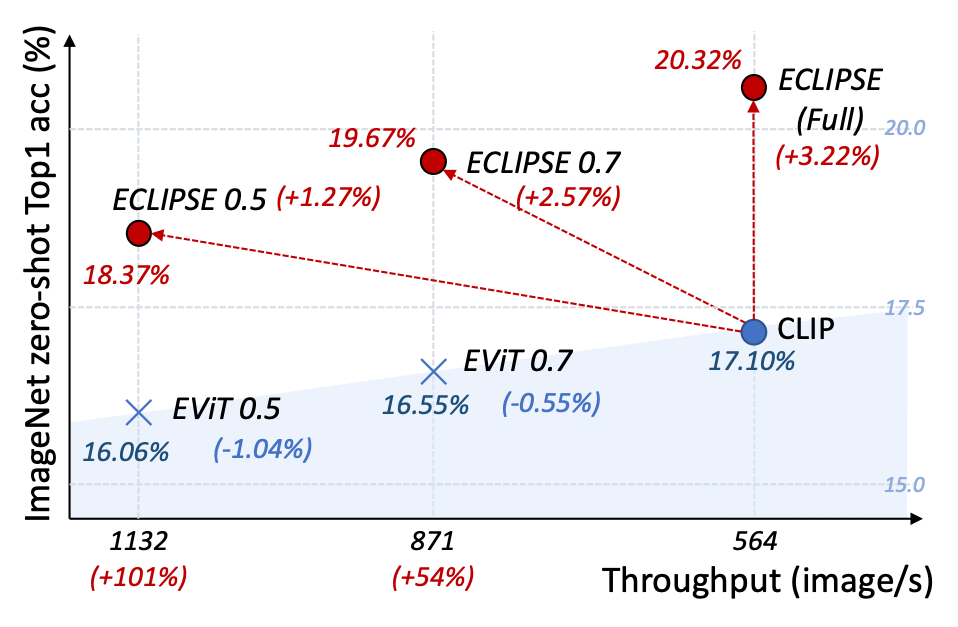}
    \caption{Time vs. ImageNet zero-shot performance analysis for Contrastive Language-Image Pretraining with existing ViT accleration framework (EViT).
    We compare the results between EViT directly applied on CLIP and EViT trained with our proposed meta-architecture, ECLIPSE.
    Our proposed framework enables even streamlined ViTs with 101\% faster throughputs to outperform the full ViT of CLIP.
    Model performance and inference time are measured with ViT-B/16 backbone.
    }
    \label{fig:fig_speed}
\end{figure}
Transformers~\cite{vaswani2017attention} have achieved significant progress across various challenging vision tasks such as image classification~\cite{dosovitskiy2021an,touvron2021training,jiang2021all,graham2021levit}, object detection~\cite{carion2020end}, semantic segmentation~\cite{xie2021segformer,liu2021swin,wang2021pyramid} and visual relationship detection~\cite{kim2021hotr,kim2022mstr}.
Following this success in vision tasks, recent studies demonstrated that large-scale vision-language pretraining (VLP)~\cite{li2019visualbert,chen2019uniter,huang2019unicoder,li2020oscar,li2020unimo,lu2019vilbert,tan2019lxmert,jia2021scaling,radford2021learning} with ViTs is scalable to large uncurated datasets and transferable to various downstream tasks.

However, the large scale image-text pairs for VLP are usually collected from the web; thus they are often noisy, i.e., having weak correlation between the image and its corresponding text description.
To alleviate the image-text misalignment problem, previous works~\cite{li2021align,Lu2022COTS} have proposed knowledge distillation framework~\cite{hinton2015distilling} with a momentum encoder for both image and text.
However, adopting two additional momentum encoders and calculating soft alignments for distillation loss inevitably increase the computational cost for training, which hinders training for large-scale VLP.

\begin{figure}
    \centering
    \includegraphics[width=0.8\columnwidth]{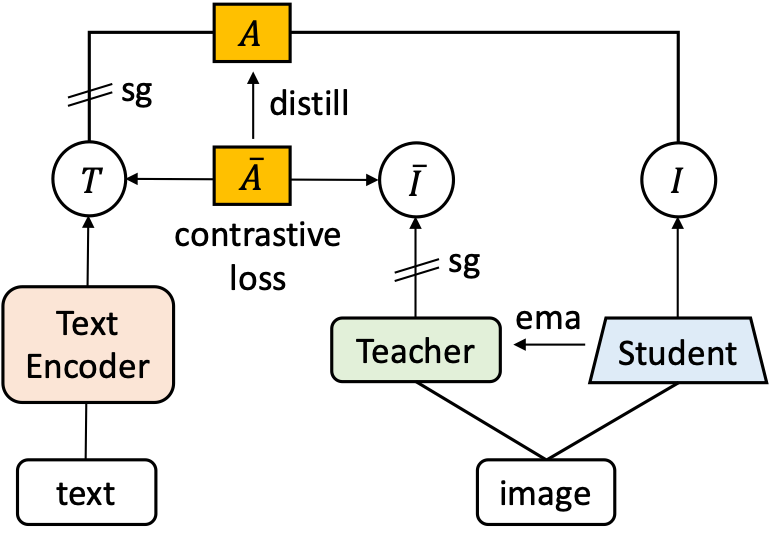}
    \vspace{0.7em}
    \caption{Overview of ECLIPSE. Student encoder is trained to estimate the soft alignment matrix $\bar{A}$ predicted by Text Encoder and the Teacher network.
    sg stands for stop-gradient, $I$ and $\bar{I}$ are encoded image with student and teacher network, respectively.
    }
    \label{fig:fig_overview}
\end{figure}

In this work, we propose a efficient formulation for distilling soft image-text alignment matrix without text momentum encoder for contrastive language-image pretraining~\cite{jia2021scaling,radford2021learning}.
Inspired from SimSiam~\cite{chen2021exploring}, we simply replace the text momentum encoder with stop-gradient operation.
This design not only eliminates the computational cost for an additional text momentum encoder, but also enables the distillation to operate within a unified projected space of text embedding, resulting in better performance.
Based on this shared projected space, we adopt token sparsification~\cite{liang2022evit} for the online image encoder to i) provide a partial view that complementarily interacts with the full-view of the momentum image encoder, ii) compensate for the computational overhead of training the momentum image encoder, and iii) accelerate inference speed.
While our distillation architecture effectively improves data efficiency by alleviating the natural misalignment between images and text, the expedited online image encoder and the momentum teacher positively interacts with a sweet spot that achieves speed improvement without degrading performance.
We name this meta-architecture as ECLIPSE: \textbf{E}xpediting \textbf{C}ontrastive \textbf{L}anguage-\textbf{I}mage \textbf{P}retraining with \textbf{S}elf-distilled \textbf{E}ncoders.
ECLIPSE is trained with a loss jointly obtained from two image-text alignment matrices (i.e., $\bar{A}$ and $A$ in Fig.~\ref{fig:fig_overview}):
\begin{itemize}
    \item The batch-wise image-text alignment matrix $\bar{A}$ between the text encoder and the momentum teacher is trained with an InfoNCE loss~\cite{oord2018representation} with hard alignment labels for matching image-text pairs.
    
    \item Student-text alignment matrix $A$ is obtained likewise with the online network and the text encoder with stop gradient. We train the online network to match $A$ with the soft alignment matrix $\bar{A}$ obtained above.
\end{itemize}
The momentum parameters are updated with an exponential moving average (EMA) of the parameters of online encoder.

Extensive experiments demonstrate the effectiveness of ECLIPSE, showing that our distillation architecture significantly improves data efficiency while achieving substantial model acceleration.
For example, when applied to CLIP~\cite{radford2021learning}, our proposed architecture improves 1.27\% zero-shot accuracy in ImageNet classification while achieving 101\% acceleration in inference speed.
Moreover, ECLIPSE can be also trained without expedition, which then shows a large 3.22\% gain compared to ViT, thus offers a model choice between an accelerated model with competitive performance and a full-capacity model with enhanced performance (see Fig.~\ref{fig:fig_speed} and Tab.~\ref{tab:keep}).
Furthermore, scaling to large-scale datasets, ECLIPSE achieves state-of-the-art on several downstream tasks, outperforming CLIP variants with a model accelerated by more than 54\%.

\section{Related Work}
Vision-Language Pretraining (VLP) learns a joint representation between two modalities on large-scale image-text pairs.
VLP covers both \textit{single}-stream models~\cite{li2019visualbert,chen2019uniter,huang2019unicoder,li2020oscar,li2020unimo,lu2019vilbert,tan2019lxmert} and \textit{dual}-stream models~\cite{jia2021scaling,radford2021learning,li2022supervision}.
Single-stream models jointly encode both image and text input with a single multi-modal encoder.
Though they have shown impressive performance in several image-text downstream tasks, single-stream models suffer from their large inference cost for the cross-modal retrieval.
Also, how to transfer the pretrained joint encoder to the unimodal downstream tasks, e.g., image recognition, is not trivial.
On the contrary, dual-stream models encode the images and texts separately with independent encoders, and thus have several advantages: simplicity, versatility, and relatively cheaper computational cost.
In this work, we focuses on a dual-stream encoder trained with a contrastive objective.

\label{sec:related_work}
\subsection{Contrastive Language-Image Pretraining}
In Contrastive Language-Image Pretraining~\cite{jia2021scaling,li2020unimo,radford2021learning}, the model is trained via a contrastive loss with large-scale image-text pairs, where the matching image-text pairs comprise a positive pair while other arbitrary pairs are treated as negative pairs.
Several works~\cite{mu2021slip,li2022supervision} introduce additional form of supervision such as self-supervision between augmented views of the image (e.g., SimCLR~\cite{chen2020simple}, SimSiam~\cite{chen2021exploring}), language self-supervision (e.g., supervision with text augmentation~\cite{wei2019eda}, masked language modeling~\cite{devlin2018bert}) and momentum contrast with nearest neighbor~\cite{he2020moco} to further improve downstream performance.
Recently, FLIP~\cite{li2022scaling} borrowed random masking strategy~\cite{he2022masked} for the input token which substantially improves the training efficiency. 
However, FLIP employs unmasked images during inference, which means that there is no speed improvement at inference time. Furthermore, due to the discrepancy between the training and test distributions, an additional unmasked tuning process is necessary.
On the other hand, ECLIPSE improves both training and inference speed without extra tuning strategy.

\subsection{Distillation and Momentum Contrast for VLP}
Knowledge distillation~\cite{hinton2015distilling} has been initially proposed to transfer the knowledge of a large model~(the teacher) to a smaller model~(the student).
Consecutive works~\cite{Romero2015FitNet,Park2019RelKD} have been explored different distilling targets other than direct output.
Extending the concept, distillation from an identically structured model~\cite{Furlanello2018BornAgain,Bagherinezhad2018LabelRefinery} or a momentum network (teacher)~\cite{Tarvainen2017meanteacher,he2020moco,grill2020bootstrap,caron2021emerging}, whose parameters are updated with the exponential moving average (EMA) of a online network (student), have been proposed.
Momentum contrast (MoCo~\cite{he2020moco}) is the pioneering contrastive learning method for images without labels that uses momentum encoder and memory queue to increase the number of negative samples. 
Inspired from MoCo, HIT~\cite{Liu2021HiT} adopted momentum encoders and memory bank for video-text contrastive matching without distillation.
ALBEF~\cite{li2021align} and COTS~\cite{Lu2022COTS} distill soft-alignment matrix obtained from both image and text momentum encoders to the online encoders.
Andonian et al.~\cite{Andonian2022robust} proposed self-distillation via swapping image-text alignment matrix without momentum encoder.
MCD~\cite{kim2023misalign} proposes a distillation where the misalignments caused by image augmentation serves as a training signal.
We introduce a novel effective distillation method called ECLIPSE whose online and momentum image encoder share text encoder.
Through a systematic analysis, we validate diverse distillation designs~(Table~\ref{tab:distill}) and demonstrate effectiveness of ECLIPSE.
\begin{figure*}
    \centering
    \includegraphics[width=0.9\textwidth]{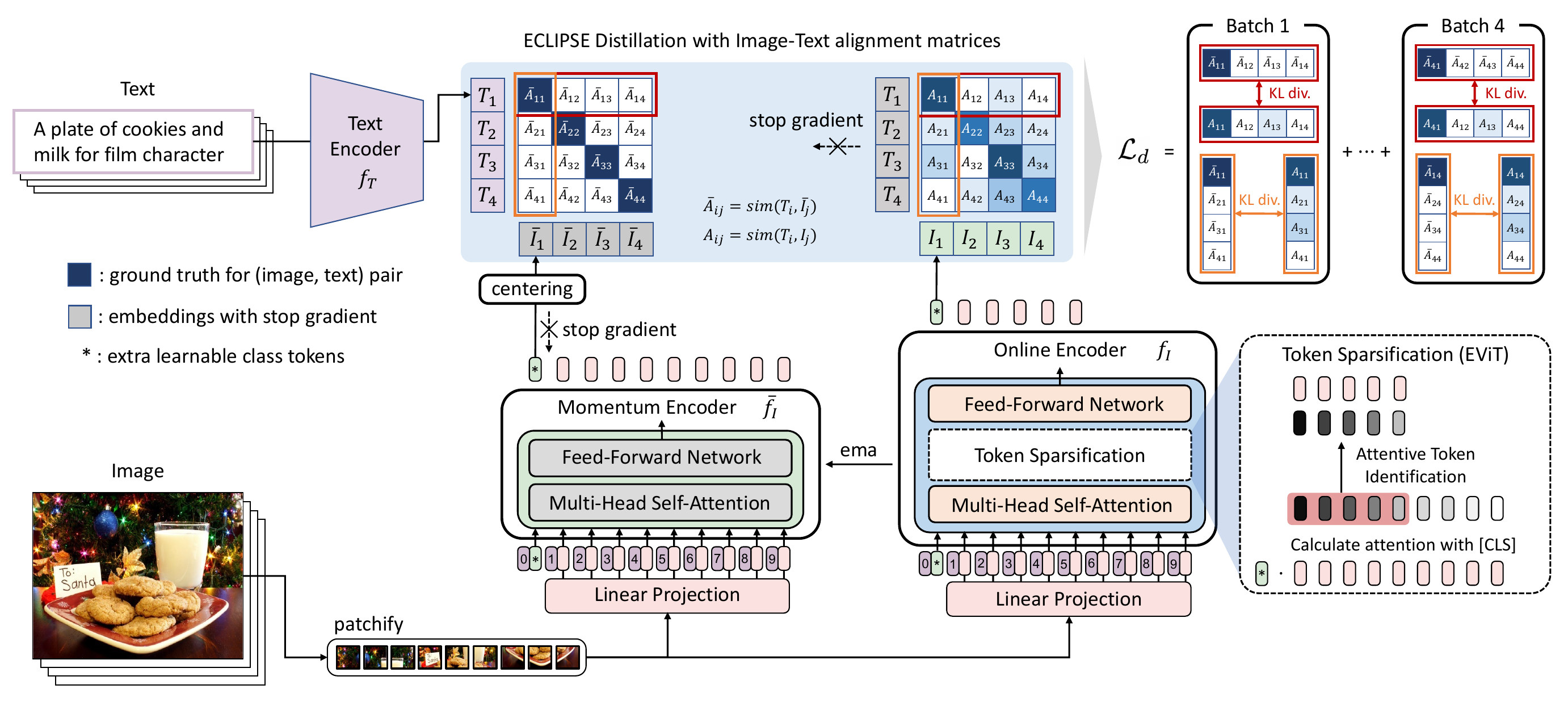}
    \caption{Overview of our proposed ECLIPSE. ECLIPSE is a meta-architecture for contrastive language-image pretraining that features a text encoder $f_T$, a momentum teacher encoder (Full ViT, $\bar{f}_I$), and a streamlined online encoder (ViT with token sparsification, $f_I$). Though the online network of ECLIPSE is compatible with any ViT acceleration method in literature~\cite{liang2022evit,rao2021dynamicvit,liang2022expediting}, we choose EViT~\cite{liang2022evit} due to its simple architecture without introducing additional parameters. Full ViTs without any sparsification can be also adopted for the online network, in which ECLIPSE then provides a full-capacity model with enhanced performance.
    }
    \label{fig:fig_architecture}
\end{figure*}
\section{Method}
\label{sec:method}

In this section, we propose ECLIPSE: Expediting Contrastive Language-Image Pretraining with Self-distilled Encoders.
Our goal is to resolve image-text misalignment problem of Contrastive Language-Image Pretraining (i.e., CLIP) for uncurated image-text pairs via efficient distillation formulation without extra text momentum encoder.
We further compensate the heavy computational cost of distillation by adopting model expediting~(i.e., EViT) to the online encoder that requires gradient computation.
We start from revisiting basic concepts of CLIP and EViT~\cite{liang2022evit}.
Then, we introduce our meta-architecture ECLIPSE that combines CLIP with our novel knowledge distillation structure and ViT acceleration for efficient training and inference.

\subsection{Contrastive Language-Image Pre-training}

First, we revisit basic form of contrastive language--image pretraining~\cite{radford2021learning}.
CLIP features a dual-encoder architecture where the image encoder $f_I$ and text encoder $f_T$ are jointly trained with contrastive objective $\mathcal{L}_{C}$.

\paragraph{Image-Text Alignment Matrix.}
For convenience, we denote $A\in\mathbb{R}^{N\times N}$ as the image-text alignment matrix for a given batch of $N$ image-text pairs $\{(x_i^I,x_i^T)\}_{i=1}^N$.
Each element of the image-text alignment matrix $A_{ij}$ is the cosine similarity between the projected representations of the $i$-th text and $j$-th image (i.e., $T_i=f_T(x^T_i)$ and $I_j=f_I(x^I_j)$, respectively), written as:
\begin{equation}
\label{eq:align}
    A_{ij}=\mbox{sim}(T_i, I_j), 
\end{equation}
where sim$(\cdot)$ is cosine similarity.

\paragraph{InfoNCE Loss.}
In CLIP, the encoded image features $I$ and text features $T$ are projected to the same dimension where the embeddings for matching image-text pairs are pulled together while embeddings for non-matched pairs are pushed apart with the InfoNCE loss~\cite{oord2018representation}.
Using Eq.~(\ref{eq:align}), the InfoNCE loss $\mathcal{L}_N$ is rewritten as:
\begin{equation}
\label{eq:infoNCE}
    \mathcal{L}_{N}(A)=-\frac{1}{N}\sum_{i=1}^N\log{\frac{\exp{\big(A_{ii}/\tau\big)}}{\sum_{j=1}^N\exp{\big(A_{ij}/\tau\big)}}},
\end{equation}
where $\tau$ is a learnable temperature variable.
The loss for the text encoder $\mathcal{L}_{T}$ and image encoder $\mathcal{L}_{I}$ are written as:
\begin{equation}
\label{eq:CLIP}
    \mathcal{L}_{T}= \mathcal{L}_{N}(A), ~~\mathcal{L}_{I}= \mathcal{L}_{N}(A^T).
\end{equation}
The overall loss for CLIP is the average of the loss for each encoder, written as $\mathcal{L}_{\text{CLIP}}(A)=\frac{1}{2}(\mathcal{L}_T+\mathcal{L}_I)$.

\subsection{Accelerating ViTs with Token Sparsification}
Previous work in ViT acceleration~\cite{rao2021dynamicvit,liang2022evit} mainly focused on token sparsification since the complexity of transformer attention is reduced at a quadratic scale with respect to the number of tokens that are discarded, significantly improving model throughputs.
Most recent works proposed token sparsification via external models or reorganizing the patch tokens based on their attentiveness with the \texttt{[CLS]} token.
In this work, we benchmark EViT~\cite{liang2022evit} since no additional parameters are introduced for acceleration.
We follow their architecture design and discard a fixed ratio ($1$-$\kappa$) of inattentive tokens according to the attention value between the \texttt{[CLS]} token and each patch in the $4^{th}$, $7^{th}$, and $10^{th}$ transformer layers, where $\kappa$ is the token keep rate.

\subsection{ECLIPSE}
Towards a data-efficient pretraining with uncurated image-text pairs, we propose ECLIPSE, a novel distillation pipeline that alleviates image-text misalignments.
The overall architecture of ECLIPSE is illustrated in Figure~\ref{fig:fig_architecture}.
ECLIPSE features a text encoder, a online image encoder (EViT), and a momentum encoder (ViT).
Below we provide a step-by-step description of our proposed ECLIPSE architecture.

\paragraph{Knowledge Distillation.}
Knowledge distillation, introduced by~\cite{hinton2015distilling}, is a learning paradigm where we train the student network to mimic the ``soft" labels predicted from the teacher network.
Following previous intuition, we adopt a knowledge distillation framework for the token-sparsified online ViT (student) to train the output of the full ViT with momentum weights (teacher), aiming to accelerate ViT without degrading performance.
However, our empirical results show that applying conventional distillation~\cite{hinton2015distilling,touvron2021training,caron2021emerging} (i.e., training the student network to directly predict the output of the teacher network) to CLIP shows minor improvement in performance when transferred to downstream tasks. 
Motivated by this finding, we propose a unique distillation architecture via the image-text alignment matrices denoted in Eq.~(\ref{eq:align}).

\paragraph{Training Loss of ECLIPSE.}
Given the momentum encoder $\bar{f}_I$, online encoder $f_I$ and text encoder $f_T$, we define a pair of alignment matrices using Eq.~(\ref{eq:align}) as:
\begin{equation}
    \bar{A}_{ij}=\mbox{sim}(T_i, \bar{I}_j), ~~A_{ij}=\mbox{sim}(\text{sg}(T_i), I_j),
\end{equation}
where $\text{sg}$ denotes stop-gradient and $\bar{I}_j=\bar{f}_I(x^I_j)$, $I_j=f_I(x^I_j)$ is the projected representations of $j$-th image with the momentum encoder and online encoder, respectively.
Note that gradient is not calculated for the momentum encoder $\bar{I}$ as it is updated by EMA of the online encoder $I$.

We first obtain the loss for the teacher-text alignment matrix $\bar{A}$ with InfoNCE loss in Eq.~(\ref{eq:infoNCE}), denoted as $\mathcal{L}_{\text{CLIP}}(\bar{A})$.
Instead of training the online network to directly predict the output of the momentum network, we distill knowledge by predicting $A$ to match $\bar{A}$.
We define the distillation loss with KL divergence for each row and column between two matrices.
Let $\sigma$ be the softmax function, the KL divergence between $A$ and $\bar{A}$ is rewritten as:
\begin{equation}
    D_{\text{KL}}(\bar A||A) = \sum_{i=1}^N \sigma(\bar A_i) \log \frac{\sigma(A_i)}{\sigma(\bar A_i)}.
\end{equation}
The overall distillation loss is the average of KL loss for row vectors and column vectors, written as $\mathcal{L}_{\text{distill}}(\bar{A},A)=\frac{1}{2}(D_{\text{KL}}(\bar A|| A) + D_{\text{KL}}(\bar A^T|| A^T))$.

To accelerate training for the online network, we balance $\mathcal{L}_{\text{distill}}$ with InfoNCE loss $\mathcal{L}_{\text{CLIP}}(A)$~\cite{touvron2021training}.
The final loss of the online network is then written as:
\begin{equation}
    \mathcal{L_{\text{online}}}=\lambda\mathcal{L}_{\text{CLIP}}(A)+(1-\lambda) \mathcal{L}_{\text{distill}}(\bar{A},A),
    \label{eq:distill}
\end{equation}
where $\lambda$ is a parameter that balances the KL divergence loss and the InfoNCE loss.
The final loss for ECLIPSE is then written as:
\begin{equation}
    \mathcal{L}=\mathcal{L}_{\text{online}}+\mathcal{L}_{\text{CLIP}}(\bar{A}).
\end{equation}

\paragraph{Momentum Update.}
Let $\theta_{f_I}$, $\theta_{\bar{f}_I}$ be the parameter of the online image encoder and momentum encoder, respectively.
For the $t$-th step, we update $\theta^{(t)}_{\bar{f}_I}$ of the momentum encoder according to the following:
\begin{equation}
    \theta_{\bar{f}_I}^{(t)} = m \theta_{\bar{f}_I}^{(t-1)} + (1-m) \theta_{f_I}^{(t)},
\end{equation}
where $m$ denotes the momentum parameter. We use $m=0.994$ in our experiments.
Momentum centering is also adopted for $\bar{f}_I$~\cite{caron2021emerging} (see our supplement for further discussion with regard to the momentum parameter and centering).
\section{Experiment}
\subsection{Implementation Details and Datasets}
For implementation details, our work is built on top of the open-source SLIP codebase~\cite{mu2021slip}\footnote{https://github.com/facebookresearch/SLIP \label{slip_github}}.
For DeCLIP~\cite{li2022supervision}, we follow the implementation details of the official code release\footnote{https://github.com/Sense-GVT/DeCLIP}.
The performance on GPU-machine runs for CLIP and SLIP follows the exact implementation details upon this codebase unless mentioned otherwise.
All of our models are pretrained in 16$\times$ A100 GPUs.
Further details can be found in the Appendix.

\paragraph{Pretraining datasets.}
To validate the effectiveness of ECLIPSE, we pretrain ECLIPSE on large-scale open-source datasets, CC (Conceptual Captions) 3M~\cite{sharma2018cc3m} and YFCC (Yahoo Flickr Creative Commons) 15M~\cite{Thomee2016YFCC100M}.
Furthermore, to show the scalability of ECLIPSE, we curate 88M image-text pairs\footnote{Details of our curated dataset will be in our supplement}.
Since the large-scale datasets (e.g., YFCC15M, 88M) feature extremely noisy text captions, intensive analysis is done with models pretrained on the relatively clean CC3M dataset.

\paragraph{Downstream datasets.}
Following CLIP~\cite{radford2021learning}, we evaluate the transferability of pretrained ECLIPSE on 11 widely used downstream datasets.
We also transfer to zero-shot Image-Text retrieval tasks on Flickr30K and MS-COCO datasets.
The evaluation settings for each dataset are consistent with CLIP as in the open-source implementation\footref{slip_github}.
See more details of downstream datasets in our supplement.

\subsection{Comparing ECLIPSE with CLIP variants}
\label{sec:cc3m_comparison}
\begin{table}[t!]
  \centering
  \small
  \begin{tabular}{l | c c c c}
    \toprule
    Method & VLC & $\text{SSL}$ & MLM & Top1(\%) \\ \hline \\[-9pt]
    (a) CLIP & \checkmark & & & 17.10 \\
    (b) CLIP w/ EViT & \checkmark & & & 16.55 \\
    (c) ECLIPSE (CLIP) & \checkmark & & & \textbf{19.67} \\ \hline
    (d) SLIP & \checkmark & \checkmark & & 22.94 \\
    (e) SLIP w/ EViT & \checkmark & \checkmark & & 21.32 \\
    (f) ECLIPSE (SLIP) & \checkmark & \checkmark & & \textbf{24.42} \\ \hline
    (h) DeCLIP & \checkmark & \checkmark & \checkmark & 25.40 \\
    (i) DeCLIP w/ EViT & \checkmark & \checkmark & \checkmark & 23.26 \\
    (g) \textbf{ECLIPSE} & \checkmark & \checkmark & & \textbf{26.41} \\
    \bottomrule
  \end{tabular}
  \vspace{0.5em}
  \caption{
  ImageNet-1k Top 1 zero shot accuracy with models pretrained on CC3M dataset under three training configurations. Details of each configuration is denoted in Sec.~\ref{sec:cc3m_comparison}.
  ECLIPSE outperforms previous CLIP variants~\cite{radford2021learning,mu2021slip,li2022supervision} across all training configurations.
  Note that na\"ively adopting EViT to existing methods suffers from performance drop after acceleration.
  }
  \label{tab:cc3m}
\end{table}
\begin{table}[t!]
  \tabcolsep=0.17cm 
  \centering
  \small
  \begin{tabular}{l | c c c c}
    \toprule
    \multirow{2}{*}{Method} & \multicolumn{3}{c}{$\mathcal{L_{\text{online}}}$ in Eq.~\ref{eq:distill}} & \multirow{2}{*}{Top1(\%)} \\
     & $\lambda$ & $\bar{A}$ & $A$ &  \\ \hline \\[-9pt]
    CLIP & - & - & - & 17.1 \\
    ECLIPSE & 0.5 & $T\times \bar{I}$ & $\text{sg}(T)\times I$ & 19.7 \\
    \midrule
    (a) ECLIPSE & 1.0 & $T\times \bar{I}$ & $\text{sg}(T)\times I$ & 16.1 \\
    (b) ECLIPSE & 0.5$\rightarrow$ 1 & $T\times \bar{I}$ & $\text{sg}(T)\times I$ & 18.5 \\
    \midrule
    (c) Output & 0.5 & $\bar{I}$ & $I$ & 16.8 \\
    (d) Matrix & 0.5 & $\bar{T}\times \bar{I}$ & $T\times I$ & 16.3 \\
    (e) Matrix & 0.5$\rightarrow$ 1 & $\bar{T}\times \bar{I}$ & $T\times I$ & 18.6 \\
    (f) ECLIPSE + $\bar{T}$ & 0.5 & $T\times \bar{I}$ & $\bar{T}\times I$ & 18.8 \\
    (g) ECLIPSE + $\bar{T}$ & 0.5$\rightarrow$ 1 & $T\times \bar{I}$ & $\bar{T}\times I$ & 15.9 \\
    \bottomrule
  \end{tabular}
  \vspace{0.5em}
  \caption{ImageNet-1k Top-1 zero-shot accuracy for different $\mathcal{L_{\text{online}}}$ in Eq.~\ref{eq:distill}. Different $\lambda$ and ($\bar{A}$, $A$): (a-b) $\lambda$ schedules (c) distillation of output, (d-e) distillation of momentum alignment matrix and (f-g) additional use of text momentum encoder for ECLIPSE, are tested. $T\times I$: alignment matrix between text and image embeddings; overbar indicates embeddings from momentum encoder. sg: stop-gradient.}
  \label{tab:distill}
\end{table}

We first compare ECLIPSE against other state-of-the-art Contrastive Language-Image Pretraining approaches~\cite{radford2021learning,mu2021slip,li2022supervision}.
Table~\ref{tab:cc3m} shows the ImageNet zero-shot results of ECLIPSE and other CLIP variants, each grouped under identical experimental settings.
All models are pretrained on the CC3M dataset with a learning rate 5e-4 for 40 epochs\footnote{More detailed training configuration will be provided in supplement.}.
We use $\kappa$=0.7 for EViT with a ViT-B/16 backbone.
For the first group (a,b,c), we compare models that only leverage Vision-Language Contrastive learning~(VLC) between image-text pairs without any augmentation.
In the second group (d,e,f), SimCLR loss~(SSL) with two augmented image views is added to the aforementioned VLC.
In the last group (h,i,g), we compare ECLIPSE with models trained with additional text augmentation~\cite{wei2019eda}~(EDA).

In Table~\ref{tab:cc3m}, we can observe that ECLIPSE on top of existing contrastive language-image pretraining pipelines, i.e., CLIP, SLIP, and DeCLIP, outperforms its baseline by a noticeable margin even with the online EViT encoder.
On the other hand, na\"ively applying EViT to existing pretraining pipelines (denoted as w/ EViT) results in lower performance.
Note that ECLIPSE requires less training costs (see Sec.~\ref{sec:discussion}) and achieves 54\% speed up in inference time (see Table~\ref{tab:keep}).
Furthermore, (g) is a simple extension of (f) where we add additional distillation loss for the augmented views (see details in our supplement).
Even without leveraging language self-supervision (Masked Language Modeling), our streamlined ViT of ECLIPSE outperforms DeCLIP.

\begin{table}[t!]
  \centering
  \small
  \begin{tabular}{c c c c}
    \toprule
              & CLIP                        & ECLIPSE                    & Throughput \\
    Keep Rate & Top1 Acc (\%)               & Top1 Acc (\%)              & (image/s) \\ \midrule
    1.0 (=ViT)       & \textbf{17.10}              & \textbf{20.32} (+3.22) & 564 \\ \hline
    0.9       & 16.82 (-0.28) & 19.41 (+2.31) & 662 (+17\%) \\
    0.8       & 16.68 (-0.42) & 19.57 (+2.47) & 758 (+34\%) \\
    0.7       & 16.55 (-0.55) & \textbf{19.67} (+2.57) & 871 (+54\%) \\
    0.6       & 16.32 (-0.78) & 18.80 (+1.70) & 998 (+77\%) \\
    0.5       & 16.06 (-1.04)  & 18.37 (+1.27) & 1132 (+101\%) \\
    \bottomrule
  \end{tabular}
  \caption{
  ImageNet-1k Top-1 zero-shot accuracy for CLIP and ECLIPSE after expediting vision encoders with different keep ratios~\cite{liang2022evit}. All models were pretrained on CC3M dataset with a ViT-B/16 backbone.
  The relative performance difference compared to CLIP-ViT model is presented in the paranthesis.
  }
  \label{tab:keep}
\end{table}
\begin{table*}[t]
\small
\centering
    \begin{tabular}{lccccccccccccccc}
        \toprule
        Method & \shortstack{Additional\\Supervision} & \rot{Pets} & \rot{CIFAR-10} & \rot{CIFAR-100} & \rot{SUN397} & \rot{Food-101} & \rot{Flowers} & \rot{Cars} & \rot{Caltech-101} & \rot{Aircraft} & \rot{DTD} & \rot{ImageNet} & \rot{\textbf{Average}} \\\midrule
        CLIP & - & 19.4 & 62.3 & 33.6 & 40.2 & 33.7 & 6.3 & 2.1 & 55.4 & 1.4 & 16.9 & 31.3 & 27.5\\
        SLIP & S & 28.3 & 72.2 & 45.3 & 45.1 & 44.7 & 6.8 & 2.9 & 65.9 & 1.9 & 21.8 & 38.3 & 33.9\\
        \textbf{ECLIPSE} & - & \textbf{24.7} & \textbf{67.8} & \textbf{38.8} & \textbf{44.4} & \textbf{34.0} & \textbf{6.2} & \textbf{2.8} & \textbf{56.7} & \textbf{2.1} & \textbf{19.6} & \textbf{32.7} & \textbf{30.0} \\
        \textbf{ECLIPSE} & S & \textbf{31.3} & \textbf{79.5} & \textbf{46.0} & \textbf{46.4} & 42.0 & \textbf{7.2} & \textbf{3.3} & 65.8 & \textbf{2.5} & \textbf{22.5} & \textbf{39.5} & \textbf{35.1} \\
        \bottomrule
        \toprule
            & & \multicolumn{6}{c}{Image-to-text retrieval} & \multicolumn{6}{c}{Text-to-image retrieval} \\
            & & \multicolumn{3}{c}{Flickr30k} & \multicolumn{3}{c}{COCO Captions} & \multicolumn{3}{c}{Flickr30k} & \multicolumn{3}{c}{COCO Captions} \\
            Method & \shortstack{Additional\\Supervision} & R@1 & R@5 & R@10 & R@1 & R@5 & R@10 & R@1 & R@5 & R@10 & R@1 & R@5 & R@10 \\
            \midrule
            CLIP & - & 34.9	& 63.9 & 75.9 & 20.8 & 43.9 & 55.7 & 23.4 & 47.2 & 58.9 & 13.0 & 31.7 & 42.7\\
            SLIP & S & 47.8	& 76.5 & 85.9 & 27.7 & 52.6 & 63.9 & 32.3 & 58.7 & 68.8 & 18.2 & 39.2 & 51.0\\
            \textbf{ECLIPSE} & - & \textbf{42.6} & \textbf{71.4} & \textbf{83.8} & \textbf{24.9} & \textbf{50.6} & \textbf{62.4} & \textbf{28.9} & \textbf{53.0} & \textbf{64.2} & \textbf{15.1} & \textbf{35.4} & \textbf{47.0} \\ 
            \textbf{ECLIPSE} & S & \textbf{50.2} & \textbf{77.4} & \textbf{87.5} & \textbf{27.9} & \textbf{53.9} & \textbf{65.9} & \textbf{33.6} & \textbf{59.6} & \textbf{70.9} & 17.5 & \textbf{39.6} & 50.7 \\ 
        \bottomrule
    \end{tabular}
    \caption{Zero-shot image classification performance (single-modal) on 11 downstream datasets and image--text retrieval (multi-modal) on the test splits of Flickr30k and COCO Captions with models pre-trained on YFCC15M.
    Our ECLIPSE achieves competitive performance with other state-of-the-art works while resulting in 54\% acceleration.
    Additional Supervisions other than Contrastive loss for image-text pairs are abbreviated as S: SSL between Augmentations.}
    \label{result:cls}
\end{table*}
\subsection{Ablation Study}
\label{subsec:ablation}
Here, we conduct ablation studies to validate how each component of ECLIPSE contributes to the final performance.
All models in this section are pretrained in CC3M dataset with $\kappa=0.7$ unless mentioned otherwise.

\paragraph{Variables for our Distillation.}
ECLIPSE is powered by a unique knowledge distillation structure where the image-text alignment matrix obtained by the online encoder and the text encoder predicts the alignment matrix jointly estimated by the momentum encoder and the text encoder.
In Table~\ref{tab:distill}, we ablate $\lambda$ and distillation target in Eq~\ref{eq:distill}.
First, ECLIPSE can be trained with only hard labels without distillation ($\lambda=1$).
We observe that ECLIPSE outperforms (a) learning from only hard labels, validating that our distillation loss contributes to the final performance.
We also found that (b) progressively changing $\lambda$ from 0.5 to 1 is worse than our default setting.
We also checked the other extreme case, learning from only distillation ($\lambda=0$), results in training failure as expected.
Second, we compare ECLIPSE with previously proposed (c) feature-level distillation~\cite{caron2021emerging} where the student network directly predicts the output of the momentum teacher and (d-e) soft alignment matrix distillation~\cite{li2021align,Lu2022COTS} where image-text alignment matrix obtained from online encoders predicts the alignment matrix from momentum encoders for both image and text.
We also test (f-g) replacing stop-gradient of text encoder with text momentum encoder which causes increase in training time.
The result shows the supremacy of our proposed distillation over the existing distillation methods and ECLIPSE variants with additional text momentum encoder.

\paragraph{Token Keep Rate.}
Table~\ref{tab:keep} shows time\footnote{We measure throughputs (128 batch, Avg of 100 runs) with https://github.com/youweiliang/evit/blob/master/helpers.py} vs performance analysis of different keep rates ($\kappa$) for EViT~\cite{liang2022evit}.
We compare our proposed ECLIPSE with CLIP where EViT is directly applied.
In the case of CLIP, the performance degrades as $\kappa$ is lowered.
On the other hand, ECLIPSE with ($\kappa=0.7$) shows the highest performance among keep rates excluding full vision~($\kappa=1.0$).
We conjecture that the token dropping of EViT can affect contrastive learning since the partial view of the attentive tokens can be interpreted as an additional augmentation on the student network.

\subsection{Pretraining ECLIPSE on Larger Datasets}
In this section, we pretrain ECLIPSE on larger scale dataset (e.g., YFCC15M)\footnote{More details of training configuration will be provided in supplement} and evaluate its transferability in single-modal and multi-modal downstream tasks.
For simplicity, we measure the effectiveness of our ECLIPSE model with two versions: (i) ECLIPSE using only the original image-text pair ((c) in Table~\ref{tab:cc3m}) and (ii) ECLIPSE with SimCLR~\cite{chen2020simple} loss between two augmented views ((f) in Table~\ref{tab:cc3m}).

\paragraph{Zero-shot Classification.}
For single-modal experiments, we test the zero-shot classification performance on 11 downstream datasets.
Table~\ref{result:cls} shows the zero-shot classification accuracy of ECLIPSE pretrained on YFCC15M dataset and transferred to downstream classification datasets.
For the test phase, the learned text encoder $f_T$ synthesizes a zero-shot linear classifier by embedding the arbitrary categories of the test dataset.
As classes are in the form of a single word, we use prompts including the label (e.g., ``\texttt{a photo of a \{label\}}") as in CLIP~\cite{radford2021learning}.
ECLIPSE with CLIP-level supervision outperforms CLIP across all 11 datasets, while ECLIPSE with additional supervision outperforms its corresponding baseline across 9 out of 11 datasets.

\begin{table}[t]
  \centering
  \small
  \begin{tabular}{l c c c c c c c}
    \toprule
    Method & Supervision & 3M   & 15M    & 88M \\ \midrule
    CLIP                 & C   & 17.1 & 31.3   & 57.4 \\ 
    \textbf{ECLIPSE}     & C &\textbf{19.7} & \textbf{32.7} & \textbf{60.2} \\
    \bottomrule
  \end{tabular}
  \vspace{0.5em}
  \caption{
  ImageNet-1k Top 1 zero shot accuracy for vision-language pretraining on different dataset scales. Both models were pretrained with a ViT-B/16 backbone for 3M and ViT-B/32 backbone for others.
  ECLIPSE shows a consistent tendency across various scales of pretrain datasets.
  }
  \label{tab:scale}
\end{table}

\paragraph{Image--Text Retrieval}
For multi-modal evaluations, we test the zero-shot image--text retrieval on Flickr30k and COCO Captions benchmarks.
The image-text retrieval task can be split into two sub-tasks according to the target modality: image retrieval and text retrieval.
Image-text pairs are ranked according to their similarity scores.
Table~\ref{result:cls} shows the zero-shot performance for image--text retrieval tasks of ECLIPSE pretrained on YFCC15M dataset.
Our ECLIPSE outperforms its counterpart CLIP across all measures with a considerable gap.

\paragraph{Scalability}
In this section, we examine how ECLIPSE performs under various scales of pretraining dataset.
In order to emphasize the effect of our meta-architecture ECLIPSE under image-text contrastive learning, we take the most simple form of Contrastive Language-Image Pretraining without any augmentation or self-supervision.
Table~\ref{tab:scale} shows the zero-shot ImageNet Top1 accuracy of our streamlined ViT ($\kappa=0.7$) after pretraining on each CC3M, YFCC15M, and our curated 88M dataset.
Across various scales of pretraining datasets, ECLIPSE shows consistent performance improvement, thus validating the data scalability of our proposed method.
\section{Discussion}
\label{sec:discussion}

\paragraph{Training Cost.}
In Table~\ref{tab:cost}, we compare training speed of CLIP and ECLIPSE. 
The result shows that ECLIPSE can reach similar training speed of CLIP even with distillation.
Consistent with our hypothesis, removing text momentum encoder ($\bar{T}$) and introducing expedition to the online encoder ($\kappa$=0.7) substantially boost the training speed compared to the na\"ive distillation with text momentum encoder.
We also measure the average GPU memory usage during training\footnote{Tested with 128 batches per GPU}.
With our GPU machine with 16$\times$A100 GPUs, ECLIPSE ($\kappa=0.7$) shows 18912 MiB/GPU average usage, showing a negligable increase compared to CLIP w/ EViT 18604 MiB/GPU, while being sufficiently efficient than CLIP trained with full ViT, 21758 MiB/GPU.
\begin{table}[t]
  \centering
  \small
  \begin{tabular}{l|c|c c c }
    \toprule
     & CLIP &    \multicolumn{3}{c}{ECLIPSE}   \\
    Encoder & ViT & $\kappa$=1.0 + $\bar{T}$   & $\kappa$=1.0   & $\kappa$=0.7 \\ \midrule
    Train time (s/batch)  &  0.409  & 0.500 & 0.484 & 0.415  \\
    \bottomrule
  \end{tabular}
  \vspace{0.5em}
  \caption{
  Training time comparison between CLIP and ECLIPSE variants. ECLIPSE achieves comparable training speed with CLIP even with disillation by removing text momentum encoder ($\bar{T}$) and replacing full ViT online encoder to streamline ViT. 
  }
  \label{tab:cost}
\end{table}

\paragraph{Efficient Image Self-Supervision for ECLIPSE.}
Previous CLIP variants~\cite{mu2021slip,li2022supervision} have shown that incorporating self-supervised learning with augmented image views (e.g., SimCLR~\cite{chen2020simple}, SimSiam~\cite{chen2021exploring}) to the contrastive language-image pretraining can be advantageous for learning better visual representations.
These works add additional forward and backward paths and MLP layers to treat the augmented views of an image.
For example, SLIP-style image self-supervision can easily be applied to ECLIPSE as in Table~\ref{tab:cc3m}(f).
However, this requires two additional forward and backward computations, resulting in longer training time.
Towards a more efficient self-supervised training, we here incorporate image SSL with the online and momentum branches of ECLIPSE.
We introduce a shared projection head on top of the momentum and online encoder for SSL.
This strategy is analogous to MoCo~\cite{he2020moco,Chen2020mocov2} without a memory queue.
We compare this efficient version (ECLIPSE-ES) with SLIP-style SSL and investigate the effect of augmentation in Table~\ref{tab:ssl}.

First, ECLIPSE-ES (b) surpasses the original SLIP and ECLIPSE (SLIP) even with fewer augmented views.
From this result, we assume that the momentum encoder plays an essential role in the improved performance just as MoCo outperforms SimCLR in image SSL.
Moreover, as the forward path of ECLIPSE is computed with the momentum encoder which does not require backward computation, ECLIPSE-ES features much shorter training time (see TPS column in Table~\ref{tab:ssl}).
Second, we found that feeding augmented image views for both paths is better than using the original image for the teacher encoder: ECLIPSE-ES(b) vs ECLIPSE-ES(a).
Consequently, ECLIPSE-ES demonstrates its training efficiency and opens up the possibility of advancement in integrating image SSL into VLP.

\begin{table}[t!]
    \tabcolsep=0.1cm
  \centering
  \footnotesize
  \begin{tabular}{l | c c c c c c c c }
    \toprule
    Method  & M & \multicolumn{3}{c}{Online Encoder} & Top1 & TPS\\ 
            &  & \textit{view} 1 & \textit{view} 2 & \textit{view} 3 & Acc. (\%)& (img/s)\\ \midrule
    CLIP & - & Img & - & - & 17.10 & \textbf{243} \\
    SLIP & - & Img & Aug & Aug & 22.94 & 129\\
    ECLIPSE (SLIP) & Img & Aug & Aug & - &  24.42 & 157\\
    ECLIPSE-ES (a) & Img & Aug & -  & - & 23.91 & \textbf{221}\\
    ECLIPSE-ES (b) & Aug & Aug & - & - & \textbf{25.01} & \textbf{221}\\ 
    \bottomrule
  \end{tabular}
  \caption{
  ImageNet-1k Top 1 zero shot accuracy with models pretrained on CC3M dataset under different number of image \textit{views} with either simple cropping (Img) or data augmentation (Aug). 
  M:momentum, TPS: throughput per second.
  }
  \label{tab:ssl}
\end{table}
\section{Conclusion}
\label{sec:conclusion}
We propose ECLIPSE, a meta-architecture for streamlining vision transformers under visual--language pretraining.
Our novel distillation formulation enables data-efficient training with accelerated ViTs under Contrastive Language-Image Pretraining.
ECLIPSE can be trained with or without model acceleration, thus offering a model choice between efficiency and performance.
Future works will include extending ECLIPSE frameworks to other modalities.

\bibliography{aaai24}

\end{document}